\documentclass{article}
\usepackage{spconf,amsmath,graphicx}
\usepackage[utf8]{inputenc}

\title{CEREBROVASCULAR NETWORK SEGMENTATION OF MRA IMAGES WITH DEEP LEARNING}
\name{Pedro Sanches$^{a}$\sthanks{Thanks to Capes agency for funding.}, Cyril Meyer$^{a}$, Vincent Vigon$^{b}$,  Benoît Naegel$^{a}$\sthanks{This research was founded by the Agence Nationale de la Recherche (Grant Agreement ANR12-MONU-0010)}}
\address{$^{a}$ ICube, Université de Strasbourg, CNRS, France \\
   $^{b}$Irma, Université de Strasbourg, France}

\begin{document}
%\ninept
%
\maketitle
\begin{abstract}
Deep learning has been shown to produce state of the art results in many tasks in biomedical imaging, especially in segmentation. Moreover, segmentation of the cerebrovascular structure from magnetic resonance angiography is a challenging problem because its complex geometry and topology have a large inter-patient variability. Therefore, in this work, we present a convolutional neural network approach for this problem. Particularly, a new network topology inspired by the U-net 3D and by the Inception modules, entitled Uception. In addition, a discussion about the best objective function for sparse data also guided most choices during the project. State of the art models are also implemented for a comparison purpose and final results show that the proposed architecture has the best performance in this particular context.

\end{abstract}
\section{Introduction}

Vascular diseases are a major cause of death worldwide, according to the World Health Organization, 15 million people suffer stroke worldwide each year. Moreover, the cerebrovascular system is a complex network of arteries and veins that supply the brain cells with vitally important nutrients and oxygen. In addition, the inter-individual differences of the cerebral arteries at a finer level are still not sufficiently understood \cite{Forkert2013}. 

The most popular imaging technique to address this task is the Magnetic Resonance Angiography (MRA), that consists in an MRI that takes into account the blood flow in the brain vessels when measuring, that are several methods for that, for example: Time-of-flight (TOF), phase contrast (PC) and fresh blood imaging (FBI). In this work, we used time-of-flight MRA.

One of the difficulties of the MRA data is their sparseness, meaning that they have less than 1 per cent of voxels that belong to the network. Moreover, there are some irrelevant signals, like artifacts and noise. As discussed above, some anatomical properties, like the geometry and topology, differ a lot between patients, specially in pathological cases.

Therefore, segmentation techniques are really important for helping clinicians and radiologists to interpret those images. In this work, the goal is to use deep learning approaches in order to learn a good representation of the cerebrovascular network and, therefore, segment the images as an expert would do.

In the biomedical imaging domain, these techniques have been used for classification of previously identified parts in more classes, anatomical object localization in space or time (such as organs or landmarks), image generation, image enhancement, and even combining image data with text reports. Nevertheless, segmentation tasks are still leading, in number of papers, as the most used application of deep learning in this domain \cite{Litjens2017}.

Here, we present a CNN design, entitled Uception, inspired by the U-net 3D \cite{3Dunet} and by the Inception modules \cite{Szegedy2016} for segmentation of the cerebrovascular network in MRA images. This architecture uses branches of convolutions with different kernel sizes in parallel in order to better select features in the several scales, with slower training as an drawback.

\section{Related Works}
In the field of 3D MRA imaging, several methods have been proposed to segment the vascular network: region-growing methods, differential analysis, model-based filtering, deformable models, path finding, mathematical morphology methods and hybrids between those methods \cite{Tankyevych2011}. All these methods take into account the fact that vessels are thin and elongated structures. Recently, deep learning techniques have been used to segment 3D in vivo multiphoton images of vasculature \cite{Haft2018} or retinal blood vessels \cite{Liskowski2016}.

However, the use of deep learning in the problem of vascular segmentation from 3D MRA has been only recently proposed in \cite{Tetteh2018}, where the authors used cross-hair filters to do the convolutions, demanding less memory and computing over bigger volumes. They also used some balanced cross-entropy functions and fine-tuning after an initial training with synthetic data. 

More generally, recent segmentation techniques based on deep learning use end-to-end CNNs, where there are no dense layers. One of the most notable is the U-net \cite{Ronneberger2015}, where there is an encoding path with series of convolutional layers and max-pooling layers, and a decoder path, with a series of up-sampling and convolutional layers, and between them, there are shortcut connections which make sure that the spatial information is being passed to the output. The same model was also later designed with 3D convolutions \cite{3Dunet}. 

The contracting path uses convolutions with or without padding, with pooling or convolutions with strides to shrink the features maps and decrease the dimensions of the image. Therefore, this part of the network is used as a kind of features extractor since the smaller image contains the encoded information of the input. 

Afterwards, another sequence of convolutions and up-sampling or deconvolutions in the expansive path are responsible to reconstruct the image to its original size. In order to have the spatial information for the reconstruction, the shortcut connections are made. 

Other similar networks in 3D, like the V-net, uses convolutions with strides of 2 instead of max-pooling and residual connections to improve convergence time \cite{Milletari2016}.

%As post-processing, conditional random fields were used as a form of regularization, removing false positives after a multi-scale 3D CNN \cite{Kamnitsas2017}.

\section{Data}

Our data comes from a time-of-flight MRA public dataset available in the TubeTK toolkit \cite{Bullit}, with voxel spacing of 0.5x0.5x0.8 mm with a volume size of 448x448x128 voxels. We dispose of 36 annotated images, of which 32 were used for training, 1 for validation and 3 for testing. Data come from healthy volunteers, from 18 to 74 years old, and without history of any factors likely to affect the cerebral vasculature.

In order to feed the neural network with this data a few preprocessing techniques were done. First, the images were resized to make them isotropic with a voxel size of 1mm with trilinear interpolation. Then, a clipping was done, followed by a normalization stage (dividing by the maximum intensity value in each image) in order to improve convergence. 
%It's well-known that more homogeneous data improve convergence in neural networks, so with this normalization step, all values are between 0 and 1.

\section{Methods}
		
%Here, a 3D approach was decided to be the best one since the blood vessels are barely recognizable when seeing just by 2D slices.

We introduce in this work an architecture based on the inception modules \cite{Szegedy2015} in a 3D fashion used in an U-net like model, entitled Uception. The idea behind those modules is to increase the network size in order to have a better representation. Normally, this is done by either adding convolutional layers or increasing their depth (number of channels after each layer). The downside is that when the network has a huge number of parameters, it's more prone to overfitting, especially if the number of annotated images is limited, as most cases in biomedical datasets. 
		
Therefore, the fundamental solution, that is changing from fully connected to sparsely connected architectures, was approximated by the Inception models. In those modules, the input is passed in parallel to several branches with different kernel sizes, and then, concatenated in the end. In addition, this architecture dissociate the depth information in the channels with 1D convolutions and spatial information with the 3D convolutions. Moreover, the use of different kernel sizes (5 and 7 voxels) help processing the image in two scales.
		
\begin{figure}[htb]

\begin{minipage}[t]{.48\linewidth}
  %\centering
  %\centerline{
  \includegraphics[width=4cm]{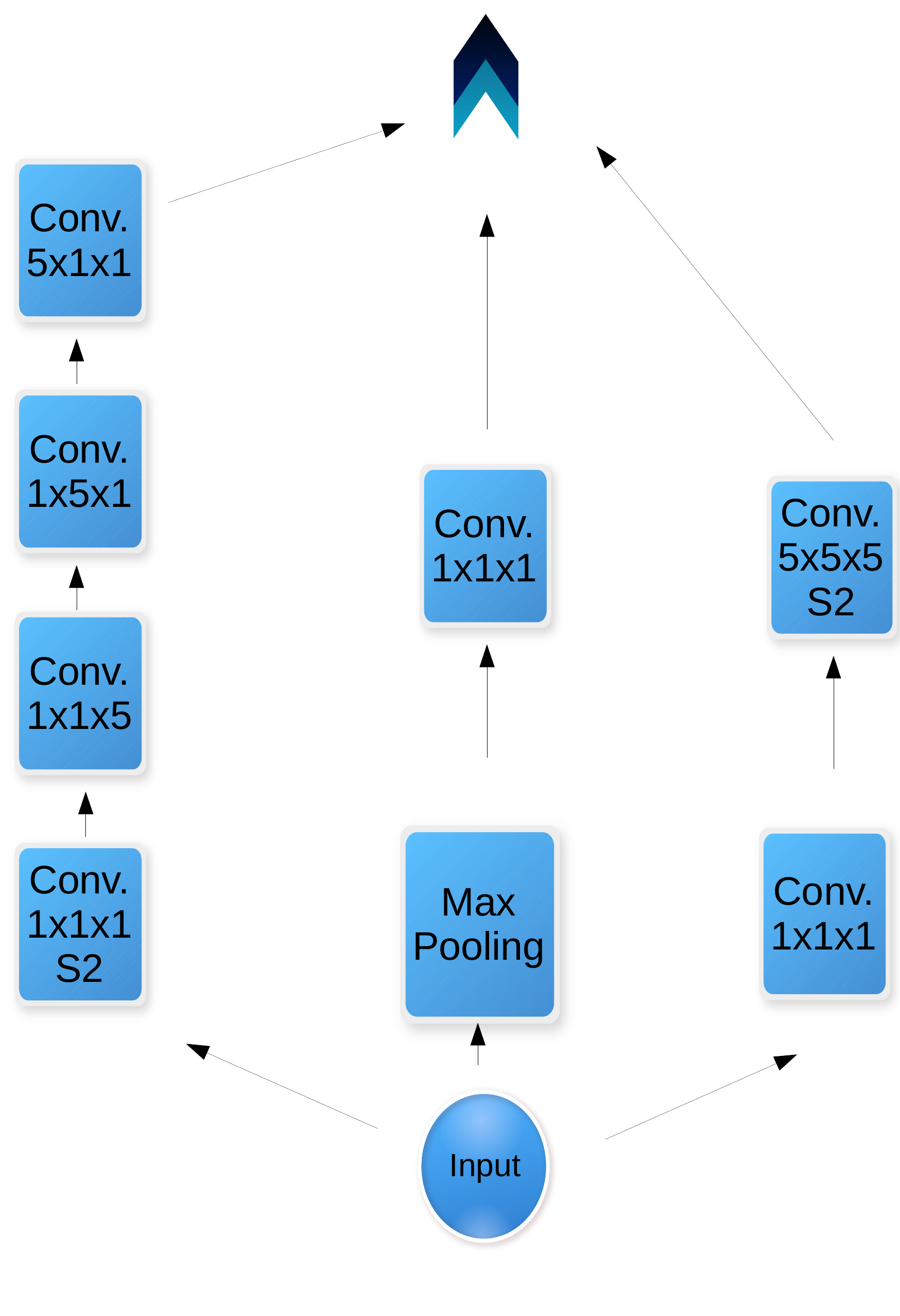}
  \centerline{(a) Reduction Block}\medskip
\end{minipage}
\begin{minipage}[t]{0.48\linewidth}
  %\centering
  %\centerline{
  \includegraphics[width=4cm]{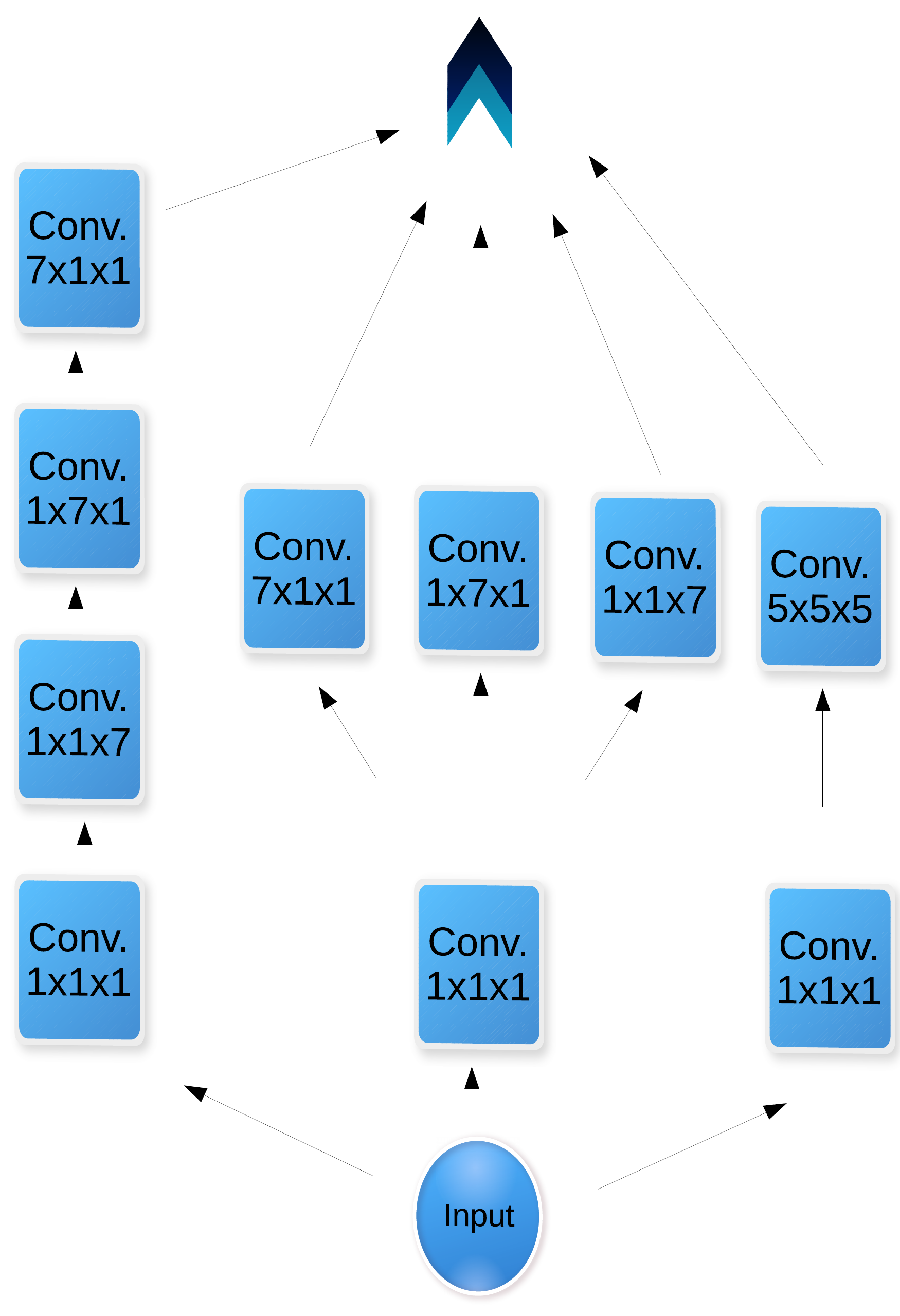}
  \centerline{(b) Deep Block}\medskip
\end{minipage}
\begin{minipage}[t]{0.9\linewidth}
%\centering
%\centerline{
\includegraphics[width=8cm]{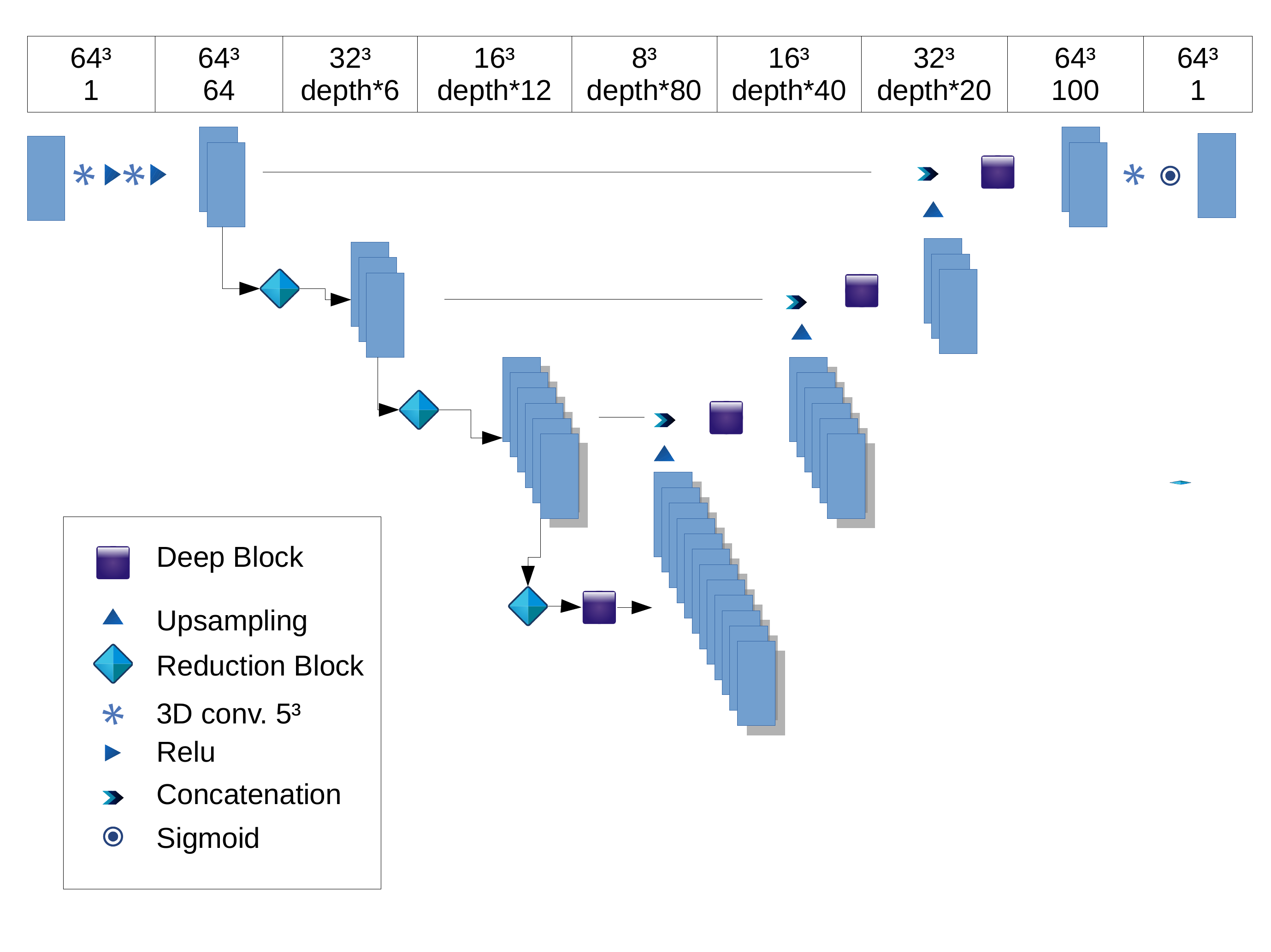}
\centerline{(c) Architecture}\medskip
\end{minipage}
\caption{Illustration of the proposed Uception architecture.}
\label{fig:uception}

\end{figure}

With ideas taken from a more recent version of the Inception V4 architecture \cite{Szegedy2016}, the following modules were created to be used in a 3D model: one model that keeps the features map shape as in fig. \ref{fig:uception}(b) and another module that halves the image size with parallel strided convolutions and max-pooling operations to make contracting path as in fig. \ref{fig:uception}(a). In the expansive path, up-sampling was used to decode the image to its original size. The final architecture is illustrated in fig. \ref{fig:uception}(c).

After each convolution layer in the hidden layers, a RELU activation function is applied. After the last convolutional layer, a Sigmoid  activation function was used. Therefore, each voxel is associated to a probability that, during the training, will approach the probability of belonging to the vessels network. Moreover, the last layer has only one channel, since the input is in binary segmentation. As regularization techniques goes, dropout was used after each activation function. 

For the loss function, we used the negative of the Dice coefficient from equation \ref{eq:dice}, where \textsc{P} is the prediction of the network and \textsc{T} is the ground truth. As its value is always between zero and one, it's numerically stable and it converged faster than other losses for this problem.
		
\begin{equation}
\label{eq:dice}
DSC(P,T) = \frac{2*|P \cap T|}{ |P|+|T|}
\end{equation}

\subsection{Training}
The training was done using the backpropagation algorithm with the Adam optimizer. Moreover, data was fed in cubic of patches 64x64x64 voxels. For the validation, all the non-superposed patches of one single image were used. In addition, the training through patches gives more data, since we don't use data augmentation and we only dispose of 36 images from this dataset.

We used a technique called snapshot ensemble \cite{Huang2017} in order to improve generalization by averaging the weights of the same model at different moments of the training. These moments are chosen as the local minima of the validation loss during a training with a cyclic learning rate schedule.

\section{Results}

\begin{figure}[ht]
\includegraphics[width=8.5cm]{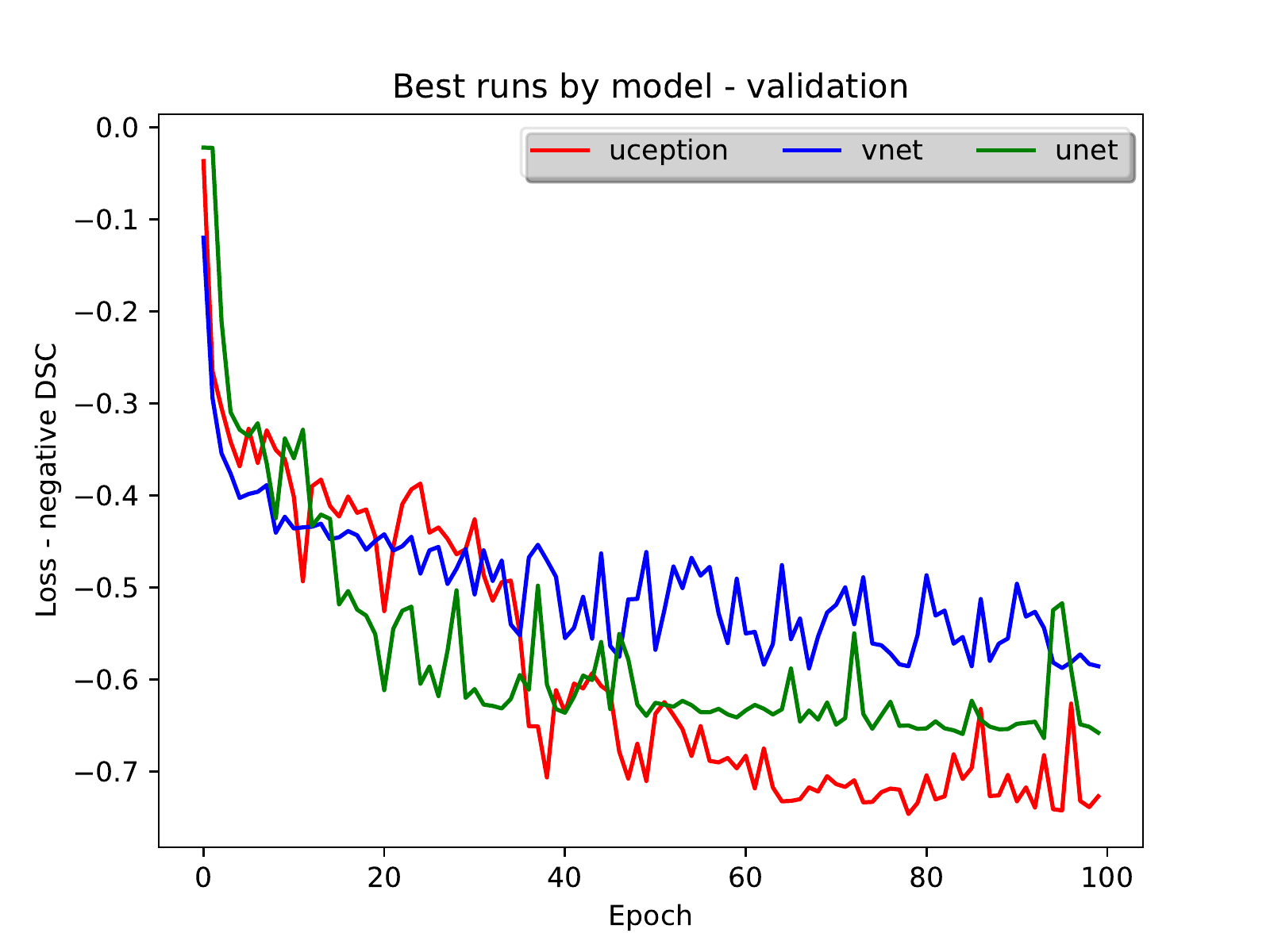}
\caption{Best Runs of the validation metrics.}
\label{fig:best_val}
\end{figure}

A hyperparameter grid search was done with the Uception and reproductions of the 3D U-net and V-net for a comparison purpose. Moreover, models with similar capacity were taken. In fig. \ref{fig:best_val}, the validation metrics during training for each architecture over 100 epochs are shown. It's clear that the proposed architecture, Uception, has less overfitting than the previous versions for segmentation of 3D brain vessels in MRA images.

For the final training, the Uception architecture was used with normal dropout at 0.25 rate, and depth equal to 10. It took around 48h to make around 1400 epochs and achieve the best results (table \ref{table:results}) over the test images \footnote{using a NVIDIA GeForce GTX 1080 GPU and TensorFlow}.  Even though these similarity metrics were measured during training and used for evaluation, we found that their significance was limited in this context, since the segmentation masks have a really good anatomical fit when superposed with the original image. Therefore, the results were also evaluated with the average Hausdorff's distance.

\begin{figure}[htb]
\begin{minipage}[b]{.48\linewidth}
  \centering
  \centerline{\includegraphics[scale=0.15]{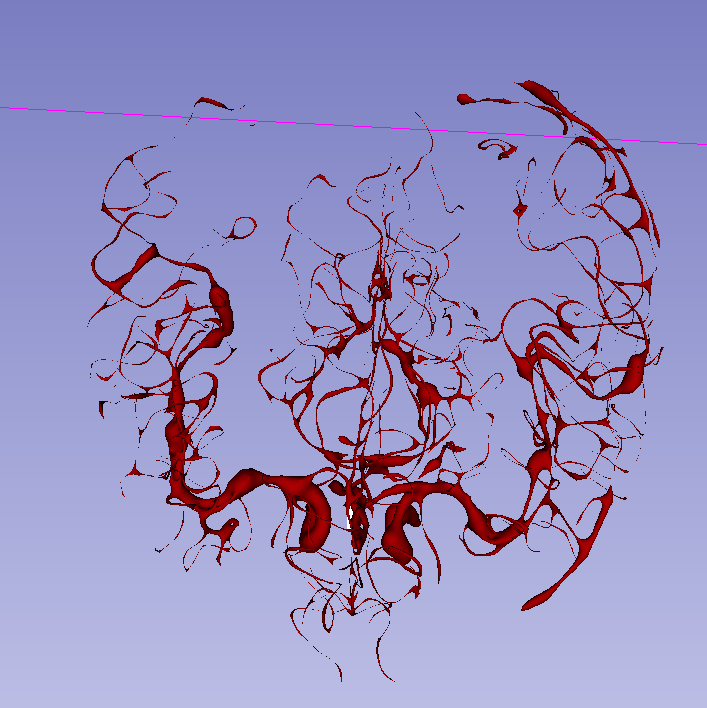}}
  \centerline{(a) Uception Segmentation}\medskip
\end{minipage}
\begin{minipage}[b]{0.48\linewidth}
  \centering
  \centerline{\includegraphics[scale=0.15]{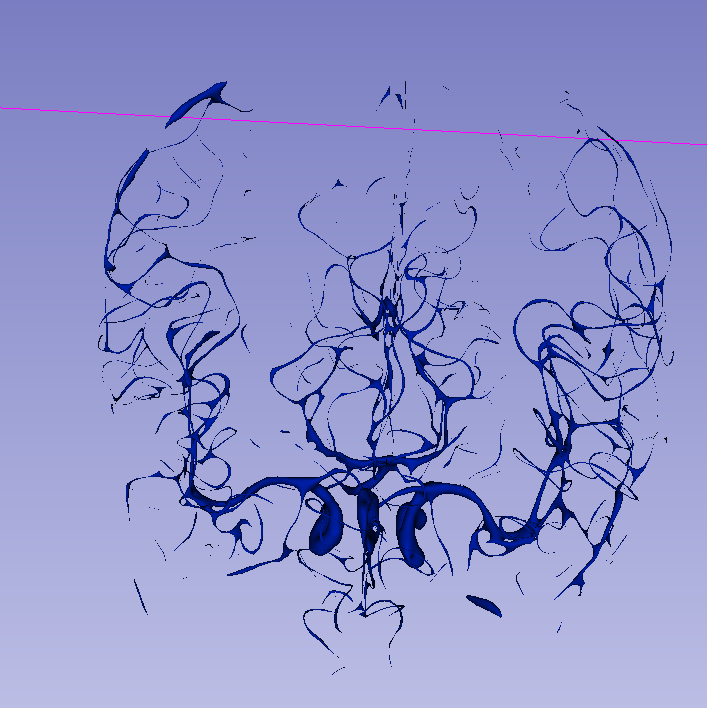}}
  \centerline{(b) Ground Truth}\medskip
\end{minipage}
\caption{Comparison between a segmentation based on Uception and the corresponding ground-truth.}
\label{fig:resukts_img}
\end{figure}

Since the sigmoid function in the last layer will give the probability that a specific voxel belongs to the cerebrovascular network, a segmentation is obtained from the network output by a thresholding at 0.9, as illustrated in fig. \ref{fig:resukts_img}(a). Moreover, the metrics are compared with a simple thresholded image at 70\% of its maximum intensity. Those were found to be the optimal thresholds for each method. Fig. \ref{fig:resukts_img}(b) represents the ground truth of the same image.

\begin{table}[ht]
\begin{tabular}{||c||c|c||} 
\hline
& Uception & Threshold \\ [0.5ex] 
\hline\hline
Dice &   0.67 $\pm$ 0.01 & 0.34$\pm$ 0.03 \\ 
\hline
Sensitivity & 0.66 $\pm$ 0.02 & 0.29 $\pm$ 0.05 \\[1ex] 
 \hline
Avg. Hausdorff Dist.[mm]  & 1.20 $\pm$  0.14 & 5.25 $\pm$  0.51 \\[1ex] 
 \hline
\end{tabular}
\caption{Results obtained for a segmentation based on Uception and a simple thresholding method.}
\label{table:results}
\end{table}

\section{Conclusion}

The goal of this work was to segment the cerebrovascular anatomical structure in magnetic resonance angiographic images using deep learning methods. Inspired by the inception architecture, we proposed an U-net like 3D architecture with blocks arranged in a more sparse manner, entitled Uception. Experiments on our dataset showed that our model has a better performance than the original U-net.

The choice of the loss function is crucial in this case due to the sparseness of the data. Normal cross-entropy functions won't converge for this problem, so optimizing the Dice coefficient appeared as a good compromise. This is due to the fact that the Dice coefficient does not take into account the total number of voxels in the data. Therefore the huge difference between the amount of voxels from the background and from the anatomical structure does not contribute to the objective function.

Evaluating these results is challenging due to the lack of precision of the available ground-truth. Moreover, the tubular structures obtained in the segmentation can be similar to those in ground-truth but with a small shift. It doesn't affect much the anatomical fidelity but really decreases Dice coefficients values. Therefore, the average Hausdorff's distance used in this case shows that our results are really close to the ground truths. 

In further works we plan to improve the results by adding in the neural network some anatomical knowledge of the cerebral vasculature. This knowledge could be based on a cerebrovascular deformable atlas, as in \cite{Dufour:2013}.

% -------------------------------------------------------------------------

% To start a new column (but not a new page) and help balance the last-page
% column length use \vfill\pagebreak.
% -------------------------------------------------------------------------

% References should be produced using the bibtex program from suitable
% BiBTeX files (here: strings, refs, manuals). The IEEEbib.bst bibliography
% style file from IEEE produces unsorted bibliography list.
% -------------------------------------------------------------------------
%\vfill\pagebreak
\bibliographystyle{IEEEbib}
\bibliography{refs}

\end{document}